\documentclass{article}

\usepackage[backend=biber, style=numeric, citestyle=numeric]{biblatex}

\usepackage[nonatbib,preprint]{neurips_2022}

\usepackage[utf8]{inputenc} 
\usepackage[T1]{fontenc}    
\usepackage{hyperref}       
\usepackage{url}            
\usepackage{booktabs}       
\usepackage{amsfonts} 
\usepackage{amsmath}
\usepackage{nicefrac}       
\usepackage{microtype}      
\usepackage{graphicx}
\usepackage{subcaption}
\usepackage{booktabs}
\usepackage{multirow}
\usepackage{colortbl}
\usepackage{tabu}
\usepackage{hhline}
\usepackage{array}
\usepackage{adjustbox}
\usepackage{boldline}
\usepackage{makecell}
\usepackage{booktabs}
\usepackage[table]{xcolor}
\usepackage{siunitx}
\usepackage{cellspace}
\usepackage{xcolor} 

\newcommand{\greenuparrow}{\textcolor{green}{\rotatebox[origin=c]{0}{$\uparrow$}}}
\newcommand{\reddownarrow}{\textcolor{red}{\rotatebox[origin=c]{0}{$\downarrow$}}}

\title{Tackling Heterogeneity in Medical Federated learning via Aligning Vision Transformers}
\author{
Erfan~Darzi$^{1,2,*}$, 
Yiqing Shen$^{3}$,
Yangming Ou$^{1,2}$,
Nanna M. Sijtsema$^{4}$, 
P.M.A van Ooijen$^{4}$\\
\\
\textit{$^{1}$ Boston Children's hospital, Boston, MA, United States}\\
\textit{$^{2}$ Harvard Medical school, Boston, MA, USA}\\
\textit{$^{3}$Johns Hopkins University, Baltimore, MD, United States}\\
\textit{$^{4}$ University Medical Center Groningen}, \textit{University of Groningen, The Netherlands}\\\\
$^{*}${Erfan.Darzi@childrens.harvard.edu}}
\begin{document}

\maketitle

\begin{abstract}
Optimization-based regularization methods have been effective in addressing the challenges posed by data heterogeneity in medical federated learning, particularly in improving the performance of underrepresented clients. However, these methods often lead to lower overall model accuracy and slower convergence rates. In this paper, we demonstrate that using Vision Transformers can substantially improve the performance of underrepresented clients without a significant trade-off in overall accuracy. This improvement is attributed to the Vision transformer's ability to capture long-range dependencies within the input data.

\end{abstract}

\section{Introduction}

Optimization-based methods have emerged as potent solutions to tackle data heterogeneity in federated setting. These methods are effective at mitigating discrepancies arising from variations in data sizes, sample numbers, or distributions across different client nodes. Despite the general effectiveness of these optimization methods, challenges specific to medical imaging in federated learning settings remain formidable.

In the realm of medical imaging, heterogeneity can manifest in a myriad of ways. These include variations in imaging modalities, different prevalence rates of specific diseases, and distinct patterns in medical datasets among hospitals. Such variations culminate in a setting of non-identical and independently distributed (non-i.i.d.) data across client nodes. This statistical heterogeneity has proven to significantly impede federated learning process. For example, heterogeneous data environments are especially challenging in specialized applications like diabetic retinopathy \cite{nasajpour2022federated}, pancreas segmentation \cite{shen2021multi}, and prostate cancer classification \cite{jiang2022harmofl}, as well as in broader contexts like bone age prediction and real-world federated brain tumor segmentation \cite{zhang2022splitavg} \cite{yue2020deep}. Such heterogeneous distributions often result in reduced diagnostic accuracy and introduce fairness concerns, particularly disadvantaging underrepresented hospitals. Addressing the complexities introduced by data heterogeneity is thus critical for the successful deployment of federated learning models in healthcare applications.


Existing federated learning methods, most notably Federated Averaging (FedAvg), face significant limitations in effectively handling heterogeneous settings \cite{zhang2022splitavg}. This has prompted various studies to explore alternative solutions that typically employ optimization techniques, such as modified training heuristics or objective functions. Techniques like SplitAvg \cite{zhang2022splitavg}, adaptive learning \cite{yeganeh2022adaptive}, hierarchical clustering \cite{gao2019hhhfl}, and proximal learning \cite{li2020federated} offer promising avenues but come with their own sets of challenges. These challenges include substantial computational complexity, the potential for overfitting due to multi-layer optimization, and constraints to specific data types. Such limitations restrict their effectiveness across a broad range of medical imaging applications. For instance, a recently proposed top-performing algorithm employs general clustering optimization in every federated round. However, it only achieves a marginal 3\% improvement over the baseline performances of FedAvg and FedAP \cite{yeganeh2022adaptive}, while demanding an order of magnitude more computational resources. This significantly complicates its practical applicability. These challenges are commonly attributed to the inherent difficulties associated with the heterogeneity problem, casting doubts on the practical utility of these models. This raises a fundamental question: do we really need to pay such a high price to mitigate the issues arising from heterogeneity?

\subsection*{Our Contributions}

We introduce the Federated Multi-Head Alignment (FedMHA) approach. This method suggests that focusing on the multi-head attention mechanism in Vision Transformers as the alignment objective can lead to improved accuracy and fairness in heterogeneous settings. The attention model's ability to handle long-range, high-dimensional distributions across diverse clients underpins this improvement. The multi-head attention mechanism's intrinsic capabilities mean that aligning it can directly affect the representation of data across clients, perhaps more than other components.

This study is driven by two main objectives. First, we address the challenges of fairness in the context of data heterogeneity in federated learning applied to medical imaging. Second, we aim to design a federated learning algorithm that achieves high accuracy levels without resorting to overly intricate optimization design to address the issue. Instead, we consider harnessing model architecture components to address heterogeneity.

Based on these objectives, our key contributions are:

\begin{itemize}
\item \textbf{Improved Fairness:} Aligning the multi-head attention mechanism in Vision Transformers between global and local models offers potential solutions to challenges posed by data heterogeneity, especially for underrepresented datasets.
\item \textbf{Enhanced Accuracy:} Our approach has consistently demonstrated superior accuracy compared to other contemporary methods. We have evaluated our model against various federated learning techniques across different levels of heterogeneity. This evaluation provides a reference for future research in federated learning for medical imaging.
\end{itemize}
\section{Problem setting and background}
\textbf{Federated learning and heterogeneity}
 Federated learning  has emerged as a decentralized approach for preserving data privacy and confidentiality while enabling models to learn from multiple data sources \cite{zhang2021subgraph}.
 In federated learning, each client owns a local private dataset ${D}_i$ drawn from distribution $\mathbb{P}_i(x,y)$, where $x$ and $y$ denote the input features and corresponding class labels, respectively. Usually, clients share a model $\mathcal{F}(\omega;x)$ with the same architecture and hyperparameters. This model is parameterized by learnable weights $\omega$ and input features $x$. The objective function of FedAvg~\cite{mcmahan2017communication} is:
\begin{equation} \label{eq:sl-lossw}
\arg\min_{\omega} \sum_{i=1}^{m} \frac{|D_i|}{N} \mathcal{L}_S(\mathcal{F}(\omega; x), y),
\end{equation}
\noindent where $\omega$ is the global model's parameters, $m$ denotes the number of clients, $N$ is the total number of instances over all clients, $\mathcal{F}$ is the shared model, and $\mathcal{L}_S$ is a general definition of any supervised learning task (e.g., a cross-entropy loss).

In a real-world FL environment, each client may represent a mobile phone with a specific user behavior pattern or a sensor deployed in a particular location, leading to statistical and/or model heterogeneous environment. In the statistical heterogeneity setting, $\mathbb{P}_i$ varies across clients, indicating heterogeneous input/output space for $x$ and $y$. For example, $\mathbb{P}_i$ on different clients can be the data distributions over different subsets of classes. In the model heterogeneity setting, $\mathcal{F}_i$ varies across clients, indicating different model architectures and hyperparameters. For the $i$-th client, the training procedure is to minimize the loss as defined below:
\begin{equation} \label{eq:fed-sp-loss}
\small
\arg\min_{\omega_1,\omega_2,\dots,\omega_m} \sum_{i=1}^{m} \frac{|D_i|}{N} \mathcal{L}_S(\mathcal{F}_i(\omega_i; x), y).
\end{equation}

Most existing methods cannot handle the heterogeneous settings above well. In particular, the fact that $\mathcal{F}_i$ has a different model architecture would cause $\omega_i$ to have a different format and size. Thus, the global model's parameter $\omega$ cannot be optimized by averaging $\omega_i$.

\textbf{Regularization in federated learning} Regularization is often employed in optimization tasks to mitigate the risk of overfitting by incorporating a penalty term to the loss function. In the context of federated learning, this is particularly useful for controlling the complexity of the global model. One popular approach is FedProx \cite{li2020federated}, which extends the FedAvg algorithm by appending a proximal term to the local optimization objective. Specifically, each client \( i \) aims to minimize:

\begin{equation}
\arg\min_{\omega_i} \mathcal{L}_S(\mathcal{F}_i(\omega_i; x), y) + \frac{\mu}{2} \left\| \omega_i - \omega \right\|^2
\end{equation}

Here, \( \mathcal{L}_S(\mathcal{F}_i(\omega_i; x), y) \) represents the local loss for client \( i \) (as defined in Eq. \ref{eq:fed-sp-loss}), \( \omega \) are the global model parameters, \( \omega_i \) are the local model parameters for client \( i \), and \( \mu \) is the regularization parameter. The server updates the global model \( \omega \) in a manner similar to FedAvg:

\begin{equation}
\omega = \sum_{i=1}^{m} \frac{|D_i|}{N} \omega_i
\end{equation}

Various other techniques like FOLB \cite{nguyen2020fast}, MOON \cite{li2021model}, and FedSplit \cite{pathak2020fedsplit} also leverage regularization to ensure that local models do not deviate significantly from the global model. However, regularization based methods like FedProx limit the global representation of models, a crucial aspect in federated learning, particularly when dealing with non-i.i.d data. This limitation stems from the predominant evaluation of these models in environments emphasizing localized structures and spatial hierarchies, primarily due to the reliance on convolution-based models. Such constraints in addressing non-i.i.d data distributions lead to the exploration of more personalized FL solutions, such as FedBN \cite{li2021fedbn}, FedPer \cite{arivazhagan2019federated}, and pFedMe \cite{Dinh2020}. 
\\FL, with its privacy-preserving capabilities, has found utility in numerous medical tasks\cite{kaissis2021end, wu2021federated, park2021federated, guo2021multi, rieke2020future}. Notable applications of FL in medical imaging are seen in multi-institutional brain tumor segmentation\cite{fedmed1, fedmed3}, breast density classification\cite{fedmed2},MRI reconstruction\cite{guo2021multi} and fMRI analysis\cite{li2020multi}.  Challenges presented by non-i.i.d. data in medical imaging, however, remain unresolved\cite{rieke2020future}, as Non-i.i.d. data largely impacts FedAvg algorithm's convergence speed\cite{li_convergence_2020, sattler_robust_2019}.

\textbf{Vision Transformers}
Dosovitskiy et al.'s Vision Transformers \cite{dosovitskiy2021image} have set benchmarks in computer vision and medical image analysis~\cite{liu2021swin,hatamizadeh2022unetr,shamshad2023transformers}. Swin Transformers~\cite{liu2021swin} enhance Vision transformers by adopting hierarchical architecture with patch merging and relative position embedding. In the medical field, Vision transformers have been intergrated in the U-shaped CNN architectures~\cite{hatamizadeh2022unetr,zhou2021nnformer}. Yet, both UNETR and nnFormer, despite their respective merits, have computational limitations due to the constraints of fixed token size and limited receptive field of CNN layers, respectively.

\section{Global-Local Encoder Alignment}

\subsection{Image representation in Vision Transformers}\label{sec:review}
The Vision Transformer \cite{vaswani2017attention,dosovitskiy2021image} is a prominent architecture for vision tasks that primarily relies on Multi-Head Self-Attention (MHSA) to model long-range dependencies among input features. Given an input tensor $\mathbf{X} \in \mathbb{R}^{H\times W\times C}$  where $H$, $W$, and $C$ are the height, width, and the feature dimension,  
 we first reshape $\mathbf{X}$  and define the query $\mathbf{Q}$, key $\mathbf{K}$, and value $\mathbf{V}$ as follows:
\begin{equation}
\begin{aligned}
\mathbf{X} \in \mathbb{R}^{H\times W\times C} &\to \mathbf{X} \in \mathbb{R}^{(H\times W)\times C}, \\
\mathbf{Q} = \mathbf{X}\mathbf{W}^q, \qquad
\mathbf{K} &= \mathbf{X}\mathbf{W}^k, \qquad
\mathbf{V} = \mathbf{X}\mathbf{W}^v,
\end{aligned}
\end{equation}
where $\mathbf{W}^q \in \mathbb{R}^{C\times C}$, $\mathbf{W}^k \in \mathbb{R}^{C\times C}$, and $\mathbf{W}^v \in \mathbb{R}^{C\times C}$ represent the linear transformation weight matrices, which are trainable. Assuming the input and output share the same dimensions, the traditional MHSA can be expressed as:
\begin{equation}\label{equ:mhsa}
\mathbf{A} = {\rm Softmax}(\mathbf{Q}\mathbf{K}^{\rm T} / \sqrt{d}) \mathbf{V},
\end{equation}
in which $\sqrt{d}$ means an approximate normalization, and the ${\rm Softmax}$ function is applied to the rows of the matrix.
We simplify the discussion by omitting the concept of multiple heads. In \ref{equ:mhsa}, the matrix product of $\mathbf{Q}\mathbf{K}^{\rm T}$ computes the pairwise similarity between tokens. Then, each new token is derived from a combination of all tokens based on their similarity.
Following the computation of  MHSA, a residual connection is added to facilitate optimization, as shown below:
\begin{equation}\label{equ:residual}
\begin{aligned}
\mathbf{X} \in \mathbb{R}^{(H\times W)\times C} &\to \mathbf{X} \in \mathbb{R}^{H\times W\times C}, \\
\mathbf{A'} &= \mathbf{A}\mathbf{W}^p + \mathbf{X},
\end{aligned}
\end{equation}
in which $\mathbf{W}^p\in \mathbb{R}^{C\times C}$ is a trainable weight matrix for feature projection.
Lastly, a multilayer perceptron (MLP) is employed to enhance the representation:
\begin{equation}\label{equ:mlp}
\mathbf{Y} = {\rm MLP}(\mathbf{A'}) + \mathbf{A'},
\end{equation}
where $\mathbf{Y}$ denotes the output of a transformer block.

It is evident that the computational complexity of MHSA (\ref{equ:mhsa}) is
\begin{equation}\label{equ:complexity}
\Omega({\rm MHSA}) = 3HWC^2 + 2H^2 W^2 C.
\end{equation}
Similarly, the space complexity (memory consumption) also includes the term of $O(H^2 W^2)$. 
As commonly known, $O(H^2 W^2)$ could become very large for high-resolution inputs.This limits the applicability of transformers for vision tasks.

\textbf{Alignment via regularization} 
The high computational complexity of MHSA as shown in equation (\ref{equ:complexity}) becomes a severe challenge in large-scale vision tasks. Moreover, in a federated learning scenario, we confront another pivotal issue: the statistical heterogeneity among different local models. Specifically, each local model may learn distinct features due to varying data distribution across clients. If not handled appropriately, this heterogeneity can hinder the global model's performance. A surrogate function could be devised that approximates the local behavior of the objective function, yet is simpler to minimize.

Let $f(x)$ represent a function. At a given point $x = y$, we can express its quadratic approximation as:

\begin{equation}
f(y) + \nabla f(y)^T(x-y) + \frac{1}{2\mu}|x-y|^2 ,
\end{equation}

where $\nabla f(y)$ is the gradient of the function $f$ at $y$ and $\mu$ is a positive scalar, representing the step size. In the context of our federated learning setting, this translates into a quadratic upper-bound for the local loss function $F_k(w)$ around the global weights $w^t$:

\begin{align}\label{equ:reg}
F_k(w) \leq F_k(w^t) + \nabla F_k(w^t)^T(w-w^t) + \frac{1}{2\mu}||\mathbf{W}^{q,k}-\mathbf{W}^{q,G}||^2_2 ,
\end{align}

where $\nabla F_k(w^t)$ is the gradient of the local loss function at $w^t$ and $\mu$ is a positive scalar representing the step size. This regularization term is analogous to the attention score in MHSA, controlling the contribution of each feature to the final representation. The norm $||\mathbf{W}^{q,k}-\mathbf{W}^{q,G}||^2_2$ represents the Euclidean distance between the local and global model's query matrices. This regularization term aligns the local model to the global model in the query space. 

The resulting objective function becomes:

\begin{align}\label{equ:modified_obj}
\min_w
h_k(w; ~w^{t}) = F_k(w) + \frac{\mu}{2}||\mathbf{W}^{q,k}-\mathbf{W}^{q,G}||^2_2,
\end{align}

where the term $||\mathbf{W}^{q,k}-\mathbf{W}^{q,G}||^2_2$ represents the squared Euclidean norm, aligning the local query matrix $\mathbf{W}^{q,k}$ to the global one $\mathbf{W}^{q,G}$. This term constrains the update of the local models, mitigating the issue of statistical heterogeneity. Here, the regularization term is analogous to the MHSA operation, where the contribution of each query (feature) to the final output depends on the similarity between the query and key. As in MHSA, where the attention weights are computed considering all tokens, here, the regularization term takes into account the whole model parameters. However, unlike MHSA, which calculates the similarity between tokens, here we calculate the distance between the local and global model's query matrices. This regularization strategy mirrors the attention mechanism in the Vision transformers. In the next section, we extend the alignment to more matrices of the vision transformers, resulting in more added terms.

\subsection{Multi-Head Encoder Alignment Mechanism (FedMHA)}


First, let's define the weight matrices of the local model $M_i$ and the global model $M_G$ as $\mathbf{W}^{q}{i}$, $\mathbf{W}^{k}{i}$, $\mathbf{W}^{v}{i}$, and $\mathbf{W}^{p}{i}$ for client $i$, and $\mathbf{W}^{q}{G}$, $\mathbf{W}^{k}{G}$, $\mathbf{W}^{v}{G}$, and $\mathbf{W}^{p}{G}$ for the global model, respectively.

Now, we can reformulate the equations (6)-(8) for each client $i$ as:

\begin{equation}
\begin{aligned}
\mathbf{Q}_i = \mathbf{X}\mathbf{W}^q_i, \qquad
\mathbf{K}_i &= \mathbf{X}\mathbf{W}^k_i, \qquad
\mathbf{V}_i = \mathbf{X}\mathbf{W}^v_i
\end{aligned}
\end{equation}


\begin{equation}
\mathbf{A}_i = {\rm Softmax}(\mathbf{Q}_i\mathbf{K}_i^{\rm T} / \sqrt{d}) \mathbf{V}_i,
\end{equation}

\begin{equation}
\begin{aligned}
\mathbf{A'}_i &= \mathbf{A}_i\mathbf{W}^p_i + \mathbf{X}_i,
\end{aligned}
\end{equation}

\begin{equation}
\mathbf{Y}_i = {\rm MLP}(\mathbf{A'}_i) + \mathbf{A'}_i,
\end{equation}

where $\mathbf{Y}_i$ denotes the output of a transformer block for the local model $M_i$.

With the MHEA method, we aim to minimize the difference between each local model encoder's weights and the global model encoder's weights. To do this, we calculate the L2 difference between each local layer's weights and the corresponding global layer's weights.

Let's denote the L2 difference between the local and global layers as $L_i^k$ for client $k$ and layer $i$. For the $Q$, $K$, and $V$ weight matrices, we compute the L2 difference as follows:

\begin{equation}
\begin{aligned}
L_{i,Q}^k &= |\mathbf{W}^{q,k}_i - \mathbf{W}^{q,G}_i|^2,
L{i,K}^k &= |\mathbf{W}^{k,k}_i - \mathbf{W}^{k,G}_i|^2,
L{i,V}^k &= |\mathbf{W}^{v,k}_i - \mathbf{W}^{v,G}_i|^2,
\end{aligned}
\end{equation}

For the MLP layers, let's denote the weight matrices as $\mathbf{W}^{MLP,k}_i$ for client $k$ and $\mathbf{W}^{MLP,G}_i$ for the global model. We compute the L2 difference for the MLP layers as follows:

\begin{equation}
L_{i,MLP}^k = |\mathbf{W}^{MLP,k}_i - \mathbf{W}^{MLP,G}_i|^2
\end{equation}

Next, we incorporate these L2 differences into the local objective function for each client $k$ and layer $i$. The modified local objective function for client $k$ and layer $i$ would be:

\begin{align}
\min_w
h_k^i(w; ~w^{t}) = F_k^i(w) + \frac{\mu}{2}\left(L_{i,Q}^k + L_{i,K}^k + L_{i,V}^k + L_{i,MLP}^k\right) ,
\end{align}

This local objective function includes both the local loss function $F_k^i(w)$ and the L2 difference between the local and global layers. The MHEA term encourages the local updates to stay close to the initial global model, addressing the issue of statistical heterogeneity and safely incorporating variable amounts of local work. The federated learning process continues for multiple rounds, with the global model sending its updated parameters to the local clients and receiving their updated parameters until a specified convergence criterion is met. 
 


\section{Data and experimental setup}

\subsection{Dataset and Pre-processing}

We utilized the IQ-OTH/NCCD Lung Cancer dataset from the Iraq-Oncology Teaching Hospital/National Center for Cancer Diseases. Collected in 2019, this dataset comprises 1190 CT scan slice images from 110 distinct instances. Each instance contains multiple slices. The CT scan images cover a window width ranging from 350 to 1200 HU and are categorized into three types: benign, malignant, and normal\cite{al2020diagnosis}.

The images are representative of a diverse patient demographic, capturing a broad spectrum of pathological conditions. For the purpose of our experiment, the dataset was partitioned across ten clients. Each client received a different number of samples, simulating a genuine federated learning environment.

For pre-processing, we standardized the images to a consistent size of 224 x 224 pixels and applied common data augmentation techniques like random rotations and horizontal flipping, as advocated in works like \cite{abunajm2023deep}. These steps align with standard practices, especially for this dataset. The choice of this dataset was influenced by its heterogeneity, with variations across gender, age, and health conditions, as noted by the original authors.

\begin{figure}[htbp]
    \centering
    \includegraphics[width=0.7\textwidth]{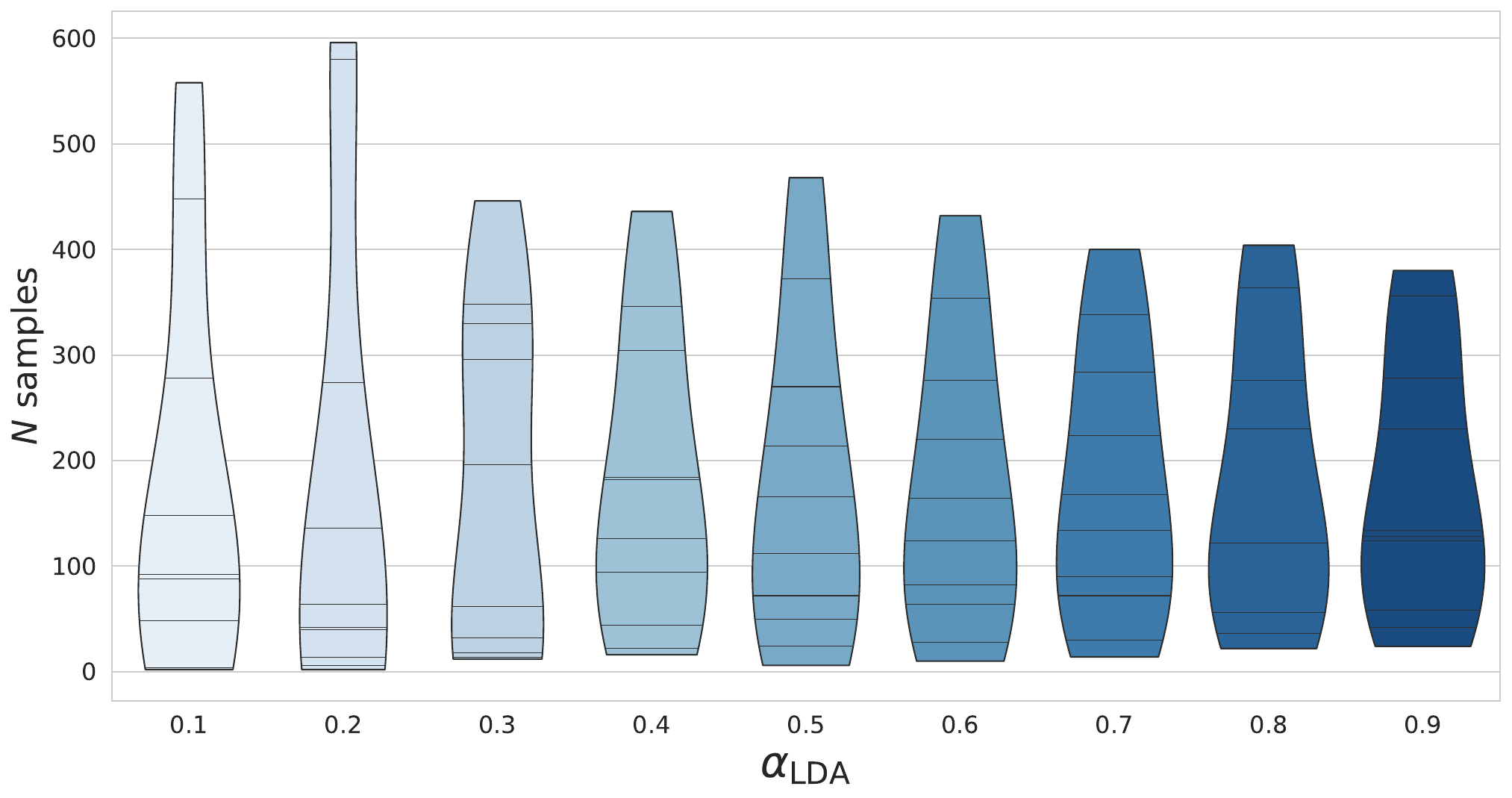}
   \caption{Client Data Distribution Variability at Different Heterogeneity Levels. The plot illustrates the variance in data distribution among clients as the heterogeneity levels, denoted by $\alpha_{\mathrm{LDA}}$ values, alter. A longer vertical axis at lower $\alpha_{\mathrm{LDA}}$ values signifies increased variability, while a wider and shorter plot at higher $\alpha_{\mathrm{LDA}}$ values suggests diminished variability.}
   \label{fig:data_distribution}
\end{figure}

\subsection{Model Architectures and Training} 

Our study compared the convolutional neural network (ConvNet5) and a pre-trained vision transformer in a federated learning setting. The ConvNet5 architecture consists of five convolutional layers, each followed by batch normalization and ReLU activation function. These are succeeded by max-pooling layers and two fully connected layers with dropout to prevent overfitting.

The dataset allocation across clients was accomplished using a Latent Dirichlet Allocation (LDA) based data splitter, which is graphically represented in Figure \ref{fig:data_distribution}. We used one A100 GPU in combination with the PyTorch framework for our experiments. We utilized the Stochastic Gradient Descent (SGD) optimizer with a learning rate of 0.01 and implemented gradient clipping with a max value of 5.0 to avoid exploding gradients. For FedProx method, the proximity coefficient, $\mu$, was set at 0.5.

\subsection{Evaluation metrics}
 The performance of the models was evaluated using metrics such as accuracy, number of correct predictions, and loss.
 For each client \( i \) with local dataset \( {D}_i \), we define the accuracy as:

\begin{equation}
\text{Acc}_{i} = \frac{1}{|{D}_i|} \sum_{x \in {D}_i} \mathbb{I}\left( y(x) = \mathcal{F}(\omega_i; x) \right)
\end{equation}

Where \( y(x) \) is the true label of instance \( x \) and \( \mathbb{I}(\cdot) \) is the indicator function, returning 1 if the model's prediction matches the true label, and 0 otherwise. Given the globally trained model, denoted by \( \mathcal{F}(\omega_{\text{global}}; x) \), the accuracy of this model on each client's test dataset \( D_{i}^{\text{test}} \) can be calculated. The worst accuracy of the global model, when evaluated across all clients, is then:
\begin{equation}
\text{Lowest Acc}_{\text{global}} = \min_{i} \left( \frac{1}{|D_{i}^{\text{test}}|} \sum_{x \in D_{i}^{\text{test}}} \mathbb{I}\left( y(x) = \mathcal{F}(\omega_{\text{global}}; x) \right) \right)
\end{equation}

This metric captures the scenario where the globally trained model has its poorest accuracy across client test datasets.
\section{Results}
\subsection{Fairness in heterogenous settings}

\begin{figure}[htbp]
\centering
\includegraphics[width=1\textwidth,viewport=0 0 880 545,clip]{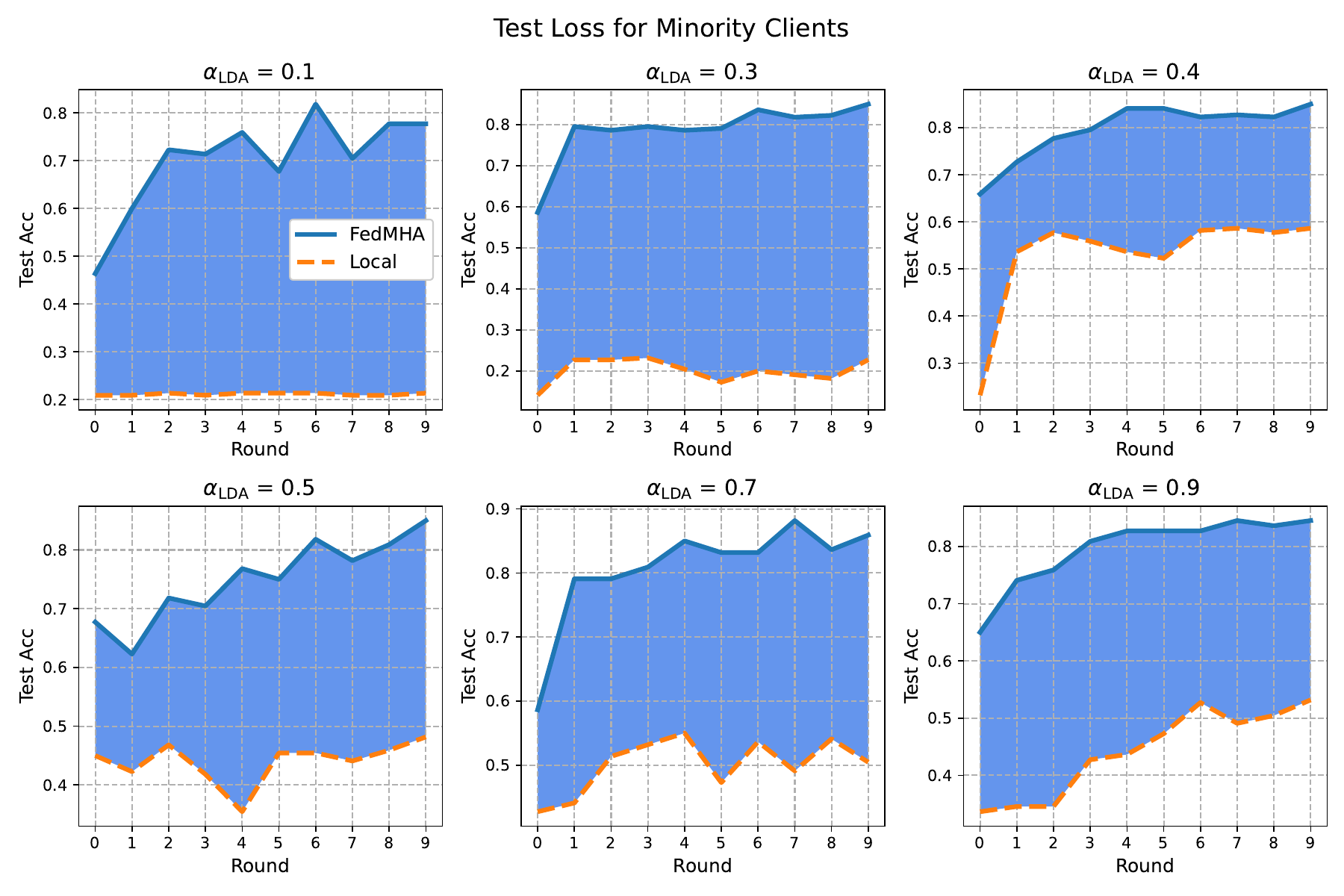}
\caption{Accuracy analysis of Multi-head encoder alignment mechanism (solid blue curves) vs. Local Stochastic Gradient Descent (SGD) (dashed orange curves) training across various heterogeneity levels. The graph shows higher accuracy improvement in higher heterogeneity levels (i.e. lower $\alpha_{\mathrm{LDA}}$)} 
\label{fig:comparison_local_sgd}
\end{figure}
We evaluate the impact of our proposed model on enhancing local models for underrepresented clients as well as for all clients in terms of test accuracy improvement over different rounds. Figure \ref{fig:comparison_local_sgd} demonstrates the difference between using a Multi-head encoder alignment mechanism (solid blue curve) and local SGD training (dashed orange curve). We observed that in highly heterogeneous settings (i.e. lower $\alpha_{\mathrm{LDA}}$ values), the improvement brought about by our model was more noticeable.The local model typically outperformed traditional settings after the first round, indicating both higher accuracy and rapid convergence rates for our approach, as depicted in the provided figures.

\begin{figure}[htbp]
\centering
\includegraphics[width=0.65\textwidth]{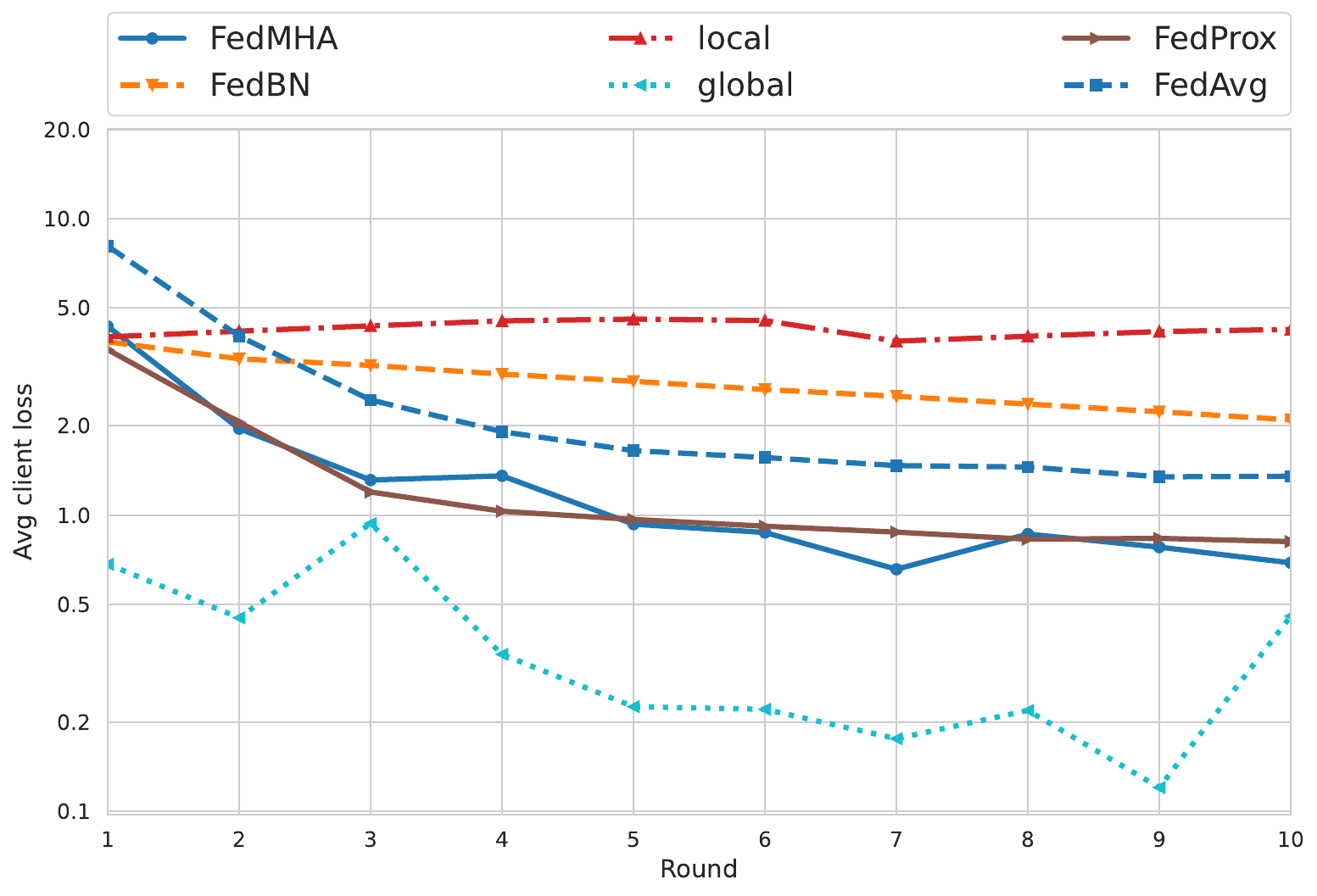}
\caption{Comparative analysis of the average loss across various federated learning settings over the initial 10 rounds. This showcases the trajectory of client loss, with the global setting employed as thebenchmark.}
\label{fig:avg_loss_comparison}
\end{figure}

To compare our proposed method with other federated learning algorithms , we trained the models for 10 rounds with LDA value of 0.2. Each method was evaluated using a cross-entropy loss function for each round. The results were then averaged based on the number of samples per client using a weighted averaging approach. Figure \ref{fig:avg_loss_comparison} provides a comparative analysis of the average loss across various federated learning settings over the initial 10 rounds. As expected, the global model with a centralized data delivers the best performance. Following the global model, FedMHA method outperforms the other federated learning algorithms. FedProx and FedAvg methods exhibit lower performance, with the FedBN approach was the least satisfactory among the considered federated learning algorithms.

\subsection{Evaluation for minority clients}

\begin{figure}[htbp]
    \centering
    \begin{adjustbox}{trim=0pt 0pt 0pt 0pt, clip, width=1.13\textwidth, center}
        \includegraphics{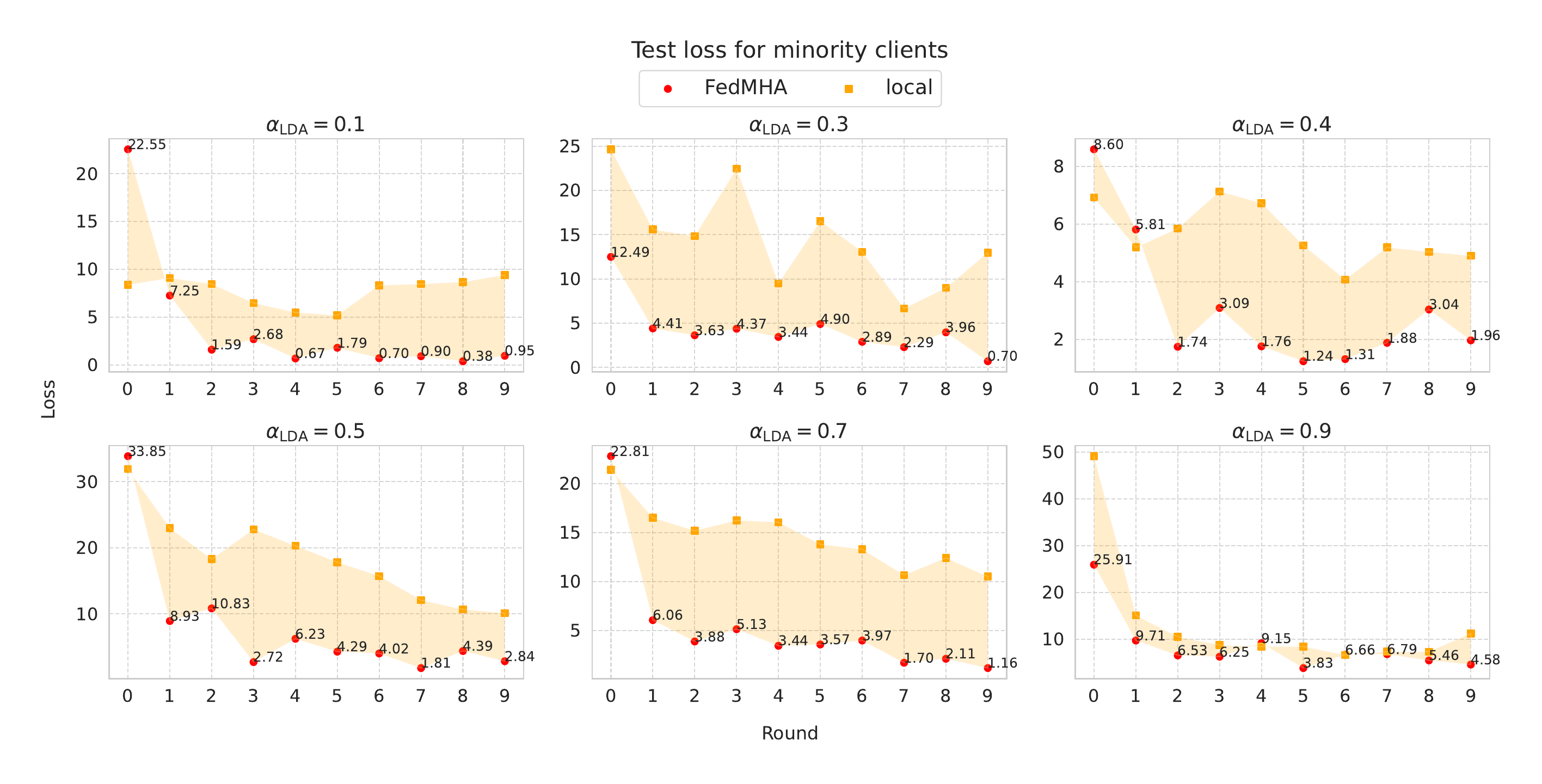}
    \end{adjustbox}
    \caption{Comparative loss analysis of our proposed Multi-head encoder alignment mechanism against Local SGD Training in a range of heterogeneity settings. Improved loss reduction is observed in highly heterogeneous environments when incorporating MHA, reflecting the effectiveness of our proposed FedMHA method.}
    \label{fig:minority_performance}
\end{figure}

In this section, we analyze the performance of minority clients in our proposed FedMHA as shown in Figure \ref{fig:minority_performance}. The purpose of this
study is to highlight the potential struggles of minority clients in heterogeneous data environments. We trained the models independently, and then evaluated them on a benchmark global dataset. Each client was trained on their own local dataset, and subsequently tested against a global dataset.

\begin{table}[ht]
\centering
\caption{Comparison of federated learning methods (Local, FedMHA, FedAvg \cite{mcmahan2017communication}, FedAvg ResNet \cite{qu2022rethinking}, FedBN \cite{li2021fedbn}, FedProx \cite{li2020federated}) under different levels of data heterogeneity represented by varying $\alpha_{\text{LDA}}$ values. The average accuracies were calculated after 5 rounds of federated learning.}
\small 
\setlength{\tabcolsep}{4pt} 
\begin{tabular}{@{}l*{7}{S[table-format=2.2,
                            table-space-text-post=\%,
                            table-column-width=12mm]}@{}}
\toprule
\rowcolor{gray!30} 
\textbf{Method} & {$\alpha_{\text{LDA}}$ = 0.1} & {$\alpha_{\text{LDA}}$ = 0.2} & {$\alpha_{\text{LDA}}$ = 0.3} & {$\alpha_{\text{LDA}}$ = 0.5} & {$\alpha_{\text{LDA}}$ = 0.7} & {$\alpha_{\text{LDA}}$ = 0.8} & {$\alpha_{\text{LDA}}$ = 0.9} \\ 
\midrule
FedAvg \cite{mcmahan2017communication} & 62.03\% & 65.55\% & \textcolor{blue}{80.24\%} & 80.03\% & 72.19\% & 85.79\% & \textcolor{blue}{84.94\%} \\ 
FedAvg (ResNet50) \cite{qu2022rethinking} & 52.73\% & 61.90\% & 69.25\% & 76.88\% & 76.06\% & 85.20\% & 84.05\% \\ 
FedBN \cite{li2021fedbn} & 47.89\% & 65.97\% & 60.93\% & 63.45\% & 61.15\% & 74.05\% & 74.10\% \\ 
FedProx \cite{li2020federated} & 47.18\% & 67.42\% & 62.99\% & 72.72\% & 71.00\% & 83.89\% & 80.43\% \\ 
\addlinespace[-1pt] 
\rowcolor[gray]{.9}
\textbf{FedMHA (ours)} & \textcolor{blue}{67.09\%} & \textcolor{blue}{74.23\%} & 70.12\% & \textcolor{blue}{81.84\%} & \textcolor{blue}{77.96\%} & \textcolor{blue}{87.76\%} & 83.99\% \\ \midrule
Local & 18.08\% & 17.85\% & 28.60\% & 46.54\% & 50.77\% & 36.67\% & 54.74\% \\ 
\bottomrule
\end{tabular}
\label{table:average_acc}
\end{table}

Table \ref{table:average_acc} provides a comparison of various federated learning methods, including our proposed FedMHA method, under different heterogeneity levels, represented by varying $\alpha_{\mathrm{LDA}}$ values. The table highlights the average accuracies achieved after 5 rounds of federated learning.A detailed analysis of these results reveals that while all models generally improve their performance as the $\alpha_{\mathrm{LDA}}$ value increases (corresponding to a more homogeneous data distribution), the FedMHA outperforms all other models, particularly in low $\alpha_{\mathrm{LDA}}$ values. 

\begin{figure}[htbp]
\centering
\begin{adjustbox}{trim=0pt 0pt 0pt 0pt, clip, width=1.1\textwidth, center}
    \includegraphics{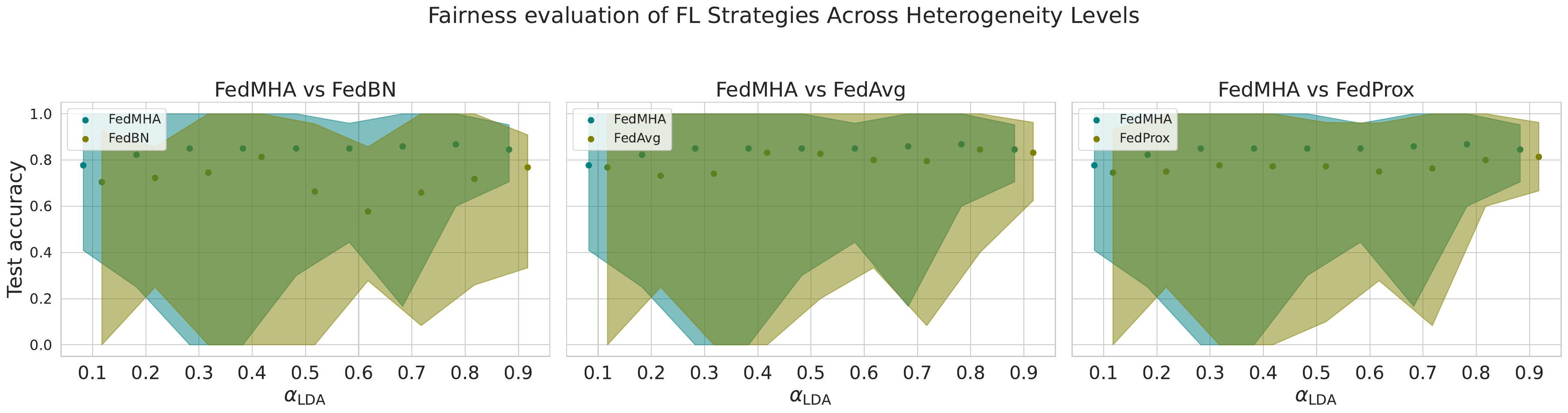}
\end{adjustbox}
\caption{Comparison of fairness in Federated Learning strategies across heterogeneity levels. Dots represent the mean accuracy for each level. The vertical stretch signifies the accuracy range for the 10 clients, with a narrower area indicating a fairer model. Our method (blue) generally outperforms other models (green), particularly in the 0.1 setting.}
\label{fig:fairness_comparison}
\end{figure}

We conduct a side-by-side comparison with other models to analyze their personlization for various models, as shown in Figure \ref{fig:fairness_comparison}. The experiments are carried out at different alpha levels and measured the average test accuracy for all clients. Our model is represented by the blue area, while the other models are depicted with a yellow area, and the overlapping area in green. Each dot in the figure represents the mean accuracy across all clients for each level of heterogeneity. The area stretching from bottom to top illustrates the range of accuracy for the ten clients involved, with a narrower area signifying a fairer model. 

The top line of our model is also higher, indicating that the performance of our method is  better in Vision Transformers for various clients, particularly in the 0.1 setting. The maximum accuracy achieved for these clients is around 0.4\%, while the minimum is close to zero. As the alpha value increases, the area becomes narrower, signifying that the personalization benefits are more pronounced for better-performing clients. To get a better understanding of the fairness, we explore the effect of three components of our training process.

\textbf{Weighted averaging boosts the effect of alignment} The first component of our investigation targets the effect of the averaging paradigm. A comparison has been made between using weighted averaging, where updates from each client are weighted by the size of their respective training sample, and a more straightforward scenario where all updates are given equal weight. The results are shown in Table \ref{table:weighted_averaging}. FedMHA shows the most noticeable enhancements when weighted averaging is employed.

 \begin{table}[ht]
\centering
\caption{Analysis of various federated learning models (FedMHA, FedAvg \cite{mcmahan2017communication}, FedAvg ResNet \cite{qu2022rethinking}, FedProx \cite{li2020federated}, FedBN \cite{li2021fedbn}, Local) with and without weighted averaging across different numbers of clients (2, 5, 8). Evaluations are made with FedAvg \cite{mcmahan2017communication} as baseline, with improvements and declines represented by \greenuparrow{} and \reddownarrow{} respectively.}
\small 
\setlength{\tabcolsep}{4pt} 
\begin{tabular}{@{}l*{6}{S[table-format=2.2,
                            table-space-text-post=\%,
                            table-column-width=14mm]}@{}}
\toprule
\rowcolor{gray!30} 
\textbf{Method} & \multicolumn{3}{c}{\textbf{W/O Weighted Averaging}} & \multicolumn{3}{c}{\textbf{With Weighted Averaging}} \\
\cmidrule(lr){2-4} \cmidrule(l){5-7}
\rowcolor{gray!30} 
& \textbf{2 Clients} & \textbf{5 Clients} & \textbf{8 Clients} & \textbf{2 Clients} & \textbf{5 Clients} & \textbf{8 Clients} \\ 
\midrule
FedAvg \cite{mcmahan2017communication}        & 63.67\% & 61.51\% & 61.70\% & 79.55\% & 76.36\% & 76.36\%  \\
FedAvg (ResNet50) \cite{qu2022rethinking}& 60.72\% \reddownarrow & 64.96\% \greenuparrow & 64.26\% \greenuparrow & 76.36\% \reddownarrow & 80.91\% \greenuparrow & 80.00\% \greenuparrow \\
FedProx\cite{li2020federated}        & 45.73\% \reddownarrow & 59.69\% \reddownarrow & 60.58\% \reddownarrow & 58.18\% \reddownarrow & 74.55\% \reddownarrow & 75.91\% \reddownarrow \\
FedBN    \cite{li2021fedbn}   & 31.12\% \reddownarrow & 54.05\% \reddownarrow & 61.35\% \reddownarrow & 38.64\% \reddownarrow & 65.45\% \reddownarrow & 74.55\% \reddownarrow 
\\
\rowcolor[gray]{.9}
\textbf{FedMHA (ours)}     & \ \ \textcolor{blue}{67.70\%} \greenuparrow & \ \ \textcolor{blue}{67.67\%} \greenuparrow & \ \ \textcolor{blue}{69.38\%} \greenuparrow & \ \  \textcolor{blue}{83.64\%} \greenuparrow & \  \ \textcolor{blue}{83.64\%} \greenuparrow & \ \ \textcolor{blue}{84.09\%} \greenuparrow \\
\midrule
Local          & 26.20\% & 26.20\% & 26.20\% & 21.36\% & 21.36\% & 21.36\% \\
\bottomrule
\label{table:weighted_averaging}
\end{tabular}
\end{table}

\textbf{Effect of Number of Clients} As the second component, we investigate the impact of the number of clients on the performance of federated learning systems. A clear correlation emerges between the number of clients and the efficacy of federated learning models. As shown in Table \ref{table:weighted_averaging}, the performance improvements associated with FedMHA, both with and without weighted averaging, span an accuracy range of 67.67\% to 84.09\%. This range contrasts the range of 21.36\% to 80.91\% for the other models.

\begin{figure}[htbp]
    \centering
 \centering
    \begin{adjustbox}{trim=0pt 0pt 0pt 0pt, clip, width=1.13\textwidth, center}
\includegraphics[viewport=0 0 1300 600,clip]{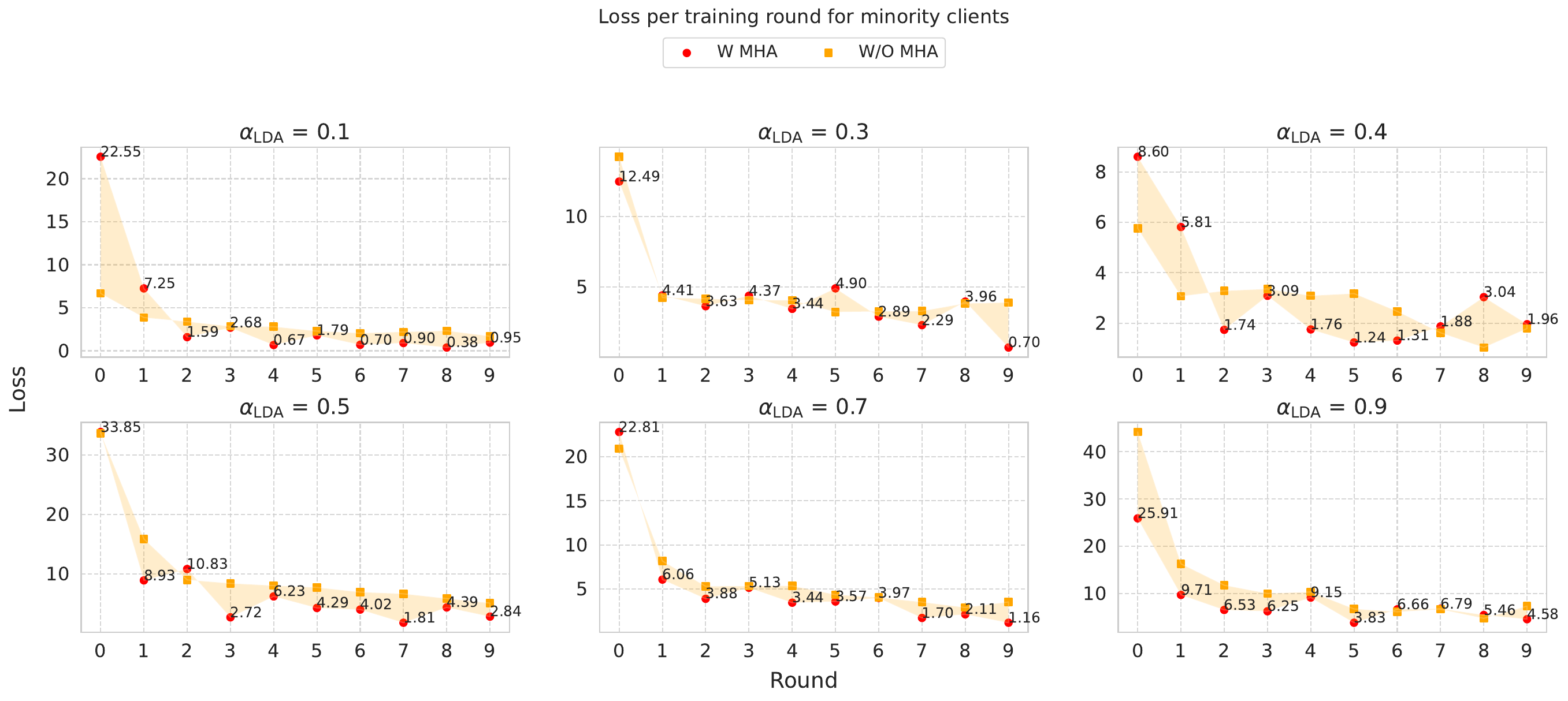}

\end{adjustbox}
    \caption{Impact of alignment loss on model performance. We compare the scenarios where the alignment term is retained in the local objective functions versus its removal. Models with the alignment term exhibit lower loss.}
\label{fig:Minority_loss_with_without_MHA}
\end{figure}
\textbf{Heterogeneity intensifies minority clients' underperformance} The third part of our investigation looks into the influence of alignment loss on the performance of federated learning models. Here, $\alpha_{\mathrm{LDA}}$ values ranging from 0.1 to 0.9 were implemented to evaluate improvements in fairness. This analysis aims to alleviate the loss experienced by worst-performing, typically underrepresented, clients. Our approach resulted in marked enhancements, especially in settings of high heterogeneity, as shown in Figure \ref{fig:Minority_loss_with_without_MHA}. Incorporating alignment loss in the local objective functions led to a boost in local training generalization and fairness for Federated Averaging.Despite achieving satisfactory performance on training data in ideal conditions, it was observed that minority clients generally underperform in settings with high levels of data heterogeneity.

Using attention layers to gather global representations from all clients and then aligning them shows great promise for improving model fairness. This likely improvement is due to the ability of attention mechanisms to effectively capture information, posing a key point of reconsideration for using convolutional layers as the main architecture in current FL algorithms \cite{li2020federated}\cite{li2021model}. This suggests a need for more focus on Vision Transformers in future updates and improvements.


\newpage
\section{Conclusion}
In this paper, we have presented and evaluated a federated learning approach that leverages Vision Transformers and multi-head attention mechanisms to effectively handle data heterogeneity in distributed settings. Our  experiments, conducted on lung cancer CT scans, demonstrate that combining optimization based approaches with vision transformer modules, outperforms existing federated learning models, particularly in scenarios with high data heterogeneity. The success of our approach in medical imaging underscores its potential in facilitating collaboration among healthcare institutions while preserving data privacy.

Our analysis  also highlights the importance of considering client data distribution and sample size during model aggregation, as a means to improve the overall accuracy. It encourages further research on employing vision transformers in heterogenous environments.The results have implications for the medical domain, where accurate diagnosis and treatment planning are paramount. Future work could focus on further enhancing fairness among clients and addressing potential scalability issues in large-scale federated learning scenarios. Additionally, exploring the use of alignment methods and vision transformers in other medical application domains could provide valuable insights into its generalizability and adaptability in the broader healthcare context. 

There are a few limitations to consider for our work. While our approach showcases the benefits of data heterogeneity handling, it doesn't address potential trade-offs related to computational overhead or communication costs. Our focus on accuracy and fairness as the primary metrics might also overlook other important aspects such as latency, or model compactness. Lastly, the real-world deployment of such algorithms may encounter challenges that are not captured in the controlled environment of our experiments.

\section*{Acknowledgement}
This research is supported by KWF Kankerbestrijding and the Netherlands
Organisation for Scientific Research (NWO)  Domain AES, as part of their joint strategic research programme: Technology for Oncology IL. The collaboration project is co-funded by the PPP allowance made available by Health Holland, Top Sector Life Sciences \& Health, to stimulate public-private partnerships. 


\printbibliography

@article{yue2020deep,
  title={Deep learning for heterogeneous medical data analysis},
  author={Yue, Lin and Tian, Dongyuan and Chen, Weitong and Han, Xuming and Yin, Minghao},
  journal={World Wide Web},
  volume={23},
  pages={2715--2737},
  year={2020},
  publisher={Springer}
}

@article{zhang2022splitavg,
  title={Splitavg: A heterogeneity-aware federated deep learning method for medical imaging},
  author={Zhang, Miao and Qu, Liangqiong and Singh, Praveer and Kalpathy-Cramer, Jayashree and Rubin, Daniel L},
  journal={IEEE Journal of Biomedical and Health Informatics},
  volume={26},
  number={9},
  pages={4635--4644},
  year={2022},
  publisher={IEEE}
}

@inproceedings{nasajpour2022federated,
  title={Federated transfer learning for diabetic retinopathy detection using CNN architectures},
  author={Nasajpour, Mohammad and Karakaya, Mahmut and Pouriyeh, Seyedamin and Parizi, Reza M},
  booktitle={SoutheastCon 2022},
  pages={655--660},
  year={2022},
  organization={IEEE}
}

@inproceedings{jiang2022harmofl,
  title={Harmofl: Harmonizing local and global drifts in federated learning on heterogeneous medical images},
  author={Jiang, Meirui and Wang, Zirui and Dou, Qi},
  booktitle={Proceedings of the AAAI Conference on Artificial Intelligence},
  volume={36},
  pages={1087--1095},
  year={2022}
}

@inproceedings{shen2021multi,
  title={Multi-task federated learning for heterogeneous pancreas segmentation},
  author={Shen, Chen and Wang, Pochuan and Roth, Holger R and Yang, Dong and Xu, Daguang and Oda, Masahiro and Wang, Weichung and Fuh, Chiou-Shann and Chen, Po-Ting and Liu, Kao-Lang and others},
  booktitle={Clinical Image-Based Procedures, Distributed and Collaborative Learning, Artificial Intelligence for Combating COVID-19 and Secure and Privacy-Preserving Machine Learning: 10th Workshop, CLIP 2021, Second Workshop, DCL 2021, First Workshop, LL-COVID19 2021, and First Workshop and Tutorial, PPML 2021, Held in Conjunction with MICCAI 2021, Strasbourg, France, September 27 and October 1, 2021, Proceedings 2},
  pages={101--110},
  year={2021},
  organization={Springer}
}

@article{pathak2020fedsplit,
  title={FedSplit: An algorithmic framework for fast federated optimization},
  author={Pathak, Reese and Wainwright, Martin J},
  journal={Advances in neural information processing systems},
  volume={33},
  pages={7057--7066},
  year={2020}
}

@article{nguyen2020fast,
  title={Fast-convergent federated learning},
  author={Nguyen, Hung T and Sehwag, Vikash and Hosseinalipour, Seyyedali and Brinton, Christopher G and Chiang, Mung and Poor, H Vincent},
  journal={IEEE Journal on Selected Areas in Communications},
  volume={39},
  number={1},
  pages={201--218},
  year={2020},
  publisher={IEEE}
}

@inproceedings{qu2022rethinking,
  title={Rethinking architecture design for tackling data heterogeneity in federated learning},
  author={Qu, Liangqiong and Zhou, Yuyin and Liang, Paul Pu and Xia, Yingda and Wang, Feifei and Adeli, Ehsan and Fei-Fei, Li and Rubin, Daniel},
  booktitle={Proceedings of the IEEE/CVF Conference on Computer Vision and Pattern Recognition},
  pages={10061--10071},
  year={2022}
}

@article{abunajm2023deep,
  title={Deep Learning Approach for Early Stage Lung Cancer Detection},
  author={Abunajm, Saleh and Elsayed, Nelly and ElSayed, Zag and Ozer, Murat},
  journal={arXiv preprint arXiv:2302.02456},
  year={2023}
}

@inproceedings{al2020diagnosis,
  title={Diagnosis of lung cancer based on CT scans using CNN},
  author={Al-Yasriy, Hamdalla F and Al-Husieny, Muayed S and Mohsen, Furat Y and Khalil, Enam A and Hassan, Zainab S},
  booktitle={IOP Conference Series: Materials Science and Engineering},
  volume={928},
  pages={022035},
  year={2020},
  organization={IOP Publishing}
}

@article{yeganeh2022adaptive,
  title={Adaptive Personlization in Federated Learning for Highly Non-iid Data},
  author={Yeganeh, Yousef and Farshad, Azade and Boschmann, Johann and Gaus, Richard and Frantzen, Maximilian and Navab, Nassir},
  journal={arXiv preprint arXiv:2207.03448},
  year={2022}
}

@article{gao2019hhhfl,
  title={Hhhfl: Hierarchical heterogeneous horizontal federated learning for electroencephalography},
  author={Gao, Dashan and Ju, Ce and Wei, Xiguang and Liu, Yang and Chen, Tianjian and Yang, Qiang},
  journal={arXiv preprint arXiv:1909.05784},
  year={2019}
}

@inproceedings{zhang2021subgraph,
  title={Subgraph federated learning with missing neighbor generation},
  author={Zhang, Ke and Yang, Carl and Li, Xiaoxiao and others},
  booktitle={Proc. Adv. Neural Inf. Process. Syst. (NIPS)},
  volume={34},
  year={2021}
}

@inproceedings{mcmahan2017communication,
  title={Communication-efficient learning of deep networks from decentralized data},
  author={McMahan, Brendan and Moore, Eider and Ramage, Daniel and others},
  booktitle={Artificial Intelligence and Statistics},
  pages={1273--1282},
  year={2017},
  organization={PMLR}
}

@inproceedings{li2020federated,
  title={Federated optimization in heterogeneous networks},
  author={Li, Tian and Sahu, Anit Kumar and Zaheer, Manzil and others},
  booktitle={Proc. Mach. Learn. Syst. (MLSys)},
  volume={2},
  pages={429--450},
  year={2020}
}

@inproceedings{li2021model,
  title={Model-contrastive federated learning},
  author={Li, Qinbin and He, Bingsheng and Song, Dawn},
  booktitle={Proc. IEEE/CVF Conf. Comput. Vis. Pattern Recognit. (CVPR)},
  pages={10713--10722},
  year={2021}
}

@article{li2021fedbn,
  title={Fedbn: Federated learning on non-iid features via local batch normalization},
  author={Li, Xiaoxiao and Jiang, Meirui and Zhang, Xiaofei and others},
  journal={arXiv preprint arXiv:2102.07623},
  year={2021}
}

@article{arivazhagan2019federated,
  title={Federated learning with personalization layers},
  author={Arivazhagan, Manoj Ghuhan and Aggarwal, Vinay and Singh, Aaditya Kumar and others},
  journal={arXiv preprint arXiv:1912.00818},
  year={2019}
}

@article{kaissis2021end,
  title={End-to-end privacy preserving deep learning on multi-institutional medical imaging},
  author={Kaissis, Georgios and Ziller, Alexander and Passerat-Palmbach, Jonathan and others},
  journal={Nat. Mach. Intell.},
  volume={3},
  number={6},
  pages={473--484},
  year={2021},
  publisher={Nature Publishing Group}
}

@inproceedings{wu2021federated,
  title={Federated Contrastive Learning for Volumetric Medical Image Segmentation},
  author={Wu, Yawen and Zeng, Dewen and Wang, Zhepeng and others},
  booktitle={Proc. Int. Conf. Med. Image Comput. Comput. Assist. Interv. (MICCAI)},
  pages={367--377},
  year={2021},
  organization={Springer}
}

@inproceedings{park2021federated,
  title={Federated Split Task-Agnostic Vision Transformer for COVID-19 CXR Diagnosis},
  author={Park, Sangjoon and Kim, Gwanghyun and Kim, Jeongsol and others},
  booktitle={Proc. Adv. Neural Inf. Process. Syst. (NIPS)},
  volume={34},
  year={2021}
}

@inproceedings{guo2021multi,
  title={Multi-institutional collaborations for improving deep learning-based magnetic resonance image reconstruction using federated learning},
  author={Guo, Pengfei and Wang, Puyang and Zhou, Jinyuan and others},
  booktitle={Proc. IEEE/CVF Conf. Comput. Vis. Pattern Recognit.},
  pages={2423--2432},
  year={2021}
}

@article{li2020multi,
  title={Multi-site fmri analysis using privacy-preserving federated learning and domain adaptation: Abide results},
  author={Li, Xiaoxiao and Gu, Yufeng and Dvornek, Nicha and Staib, Lawrence and Ventola, Pamela and Duncan, James S},
  journal={arXiv preprint arXiv:2001.05647},
  year={2020}
}

@article{fedmed3,
  title={Federated learning in medicine: facilitating multi-institutional collaborations without sharing patient data},
  author={Sheller, Micah J and Edwards, Brandon and Reina, G Anthony and Martin, Jason and Pati, Sarthak and Kotrotsou, Aikaterini and Milchenko, Mikhail and Xu, Weilin and Marcus, Daniel and Colen, Rivka R and Bakas, Spyridon},
  journal={Scientific reports},
  volume={10},
  number={1},
  pages={1--12},
  year={2020},
  publisher={Nature Publishing Group}
}

@inproceedings{hatamizadeh2022unetr,
  title={Unetr: Transformers for 3d medical image segmentation},
  author={Hatamizadeh, Ali and Tang, Yucheng and Nath, Vishwesh and Yang, Dong and Myronenko, Andriy and Landman, Bennett and Roth, Holger R and Xu, Daguang},
  booktitle={Proceedings of the IEEE/CVF winter conference on applications of computer vision},
  pages={574--584},
  year={2022}
}

@article{zhou2021nnformer,
  title={nnformer: Interleaved transformer for volumetric segmentation},
  author={Zhou, Hong-Yu and Guo, Jiansen and Zhang, Yinghao and Yu, Lequan and Wang, Liansheng and Yu, Yizhou},
  journal={arXiv preprint arXiv:2109.03201},
  year={2021}
}

@article{shamshad2023transformers,
  title={Transformers in medical imaging: A survey},
  author={Shamshad, Fahad and Khan, Salman and Zamir, Syed Waqas and Khan, Muhammad Haris and Hayat, Munawar and Khan, Fahad Shahbaz and Fu, Huazhu},
  journal={Medical Image Analysis},
  pages={102802},
  year={2023},
  publisher={Elsevier}
}

@inproceedings{liu2021swin,
  title={Swin transformer: Hierarchical vision transformer using shifted windows},
  author={Liu, Ze and Lin, Yutong and Cao, Yue and Hu, Han and Wei, Yixuan and Zhang, Zheng and Lin, Stephen and Guo, Baining},
  booktitle={Proceedings of the IEEE/CVF international conference on computer vision},
  pages={10012--10022},
  year={2021}
}

@incollection{fedmed2,
  title={Federated Learning for Breast Density Classification: A Real-World Implementation},
  author={Roth, Holger R and Chang, Ken and Singh, Praveer and Neumark, Nir and Li, Wenqi and Gupta, Vikash and Gupta, Sharut and Qu, Liangqiong and Ihsani, Alvin and Bizzo, Bernardo C and others},
  booktitle={Domain Adaptation and Representation Transfer, and Distributed and Collaborative Learning},
  pages={181--191},
  year={2020},
  publisher={Springer}
}

@article{dosovitskiy2021image,
  title={An image is worth 16x16 words: Transformers for image recognition at scale},
  author={Dosovitskiy, Alexey and Beyer, Lucas and Kolesnikov, Alexander and Weissenborn, Dirk and Zhai, Xiaohua and Unterthiner, Thomas and Dehghani, Mostafa and Minderer, Matthias and Heigold, Georg and Gelly, Sylvain and others},
  journal={arXiv preprint arXiv:2010.11929},
  year={2020}
}

@article{vaswani2017attention,
  title={Attention is all you need},
  author={Vaswani, Ashish and Shazeer, Noam and Parmar, Niki and Uszkoreit, Jakob and Jones, Llion and Gomez, Aidan N and Kaiser, {\L}ukasz and Polosukhin, Illia},
  journal={Advances in neural information processing systems},
  volume={30},
  year={2017}
}

@inproceedings{fedmed1,
  title={Privacy-preserving federated brain tumour segmentation},
  author={Li, Wenqi and Milletar{\`\i}, Fausto and Xu, Daguang and Rieke, Nicola and Hancox, Jonny and Zhu, Wentao and Baust, Maximilian and Cheng, Yan and Ourselin, S{\'e}bastien and Cardoso, M Jorge and others},
  booktitle={International Workshop on Machine Learning in Medical Imaging},
  pages={133--141},
  year={2019},
  organization={Springer}
}

@article{rieke2020future,
  title={The future of digital health with federated learning},
  author={Rieke, Nicola and Hancox, Jonny and Li, Wenqi and Milletari, Fausto and Roth, Holger and Albarqouni, Shadi and Bakas, Spyridon and Galtier, Mathieu N and Landman, Bennett and Maier-Hein, Klaus and others},
  journal={npj Digit. Med.},
  volume={3},
  pages={119},
  year={2020}
}

@String(CVPR  = {CVPR})

@String(NIPS  = {NeurIPS})

@String(AAAI = {AAAI})

@String(CVPR= {IEEE Conf. Comput. Vis. Pattern Recog.})

@String(NIPS= {Adv. Neural Inform. Process. Syst.})

@article{Dinh2020,
  title={Personalized Federated Learning with Moreau Envelopes},
  author={Canh T. Dinh and Nguyen H. Tran and Tuan Dung Nguyen},
  journal={NeurIPS},
  year={2020}
}

@article{li_convergence_2020,
	title = {On the {Convergence} of {FedAvg} on {Non}-{IID} {Data}},
	url = {http://arxiv.org/abs/1907.02189},
	urldate = {2020-12-06},
	journal = {arXiv:1907.02189 [cs, math, stat]},
	author = {Li, Xiang and Huang, Kaixuan and Yang, Wenhao and Wang, Shusen and Zhang, Zhihua},
	month = jun,
	year = {2020},
	note = {arXiv: 1907.02189},
}

@article{sattler_robust_2019,
	title = {Robust and {Communication}-{Efficient} {Federated} {Learning} from {Non}-{IID} {Data}},
	url = {http://arxiv.org/abs/1903.02891},
	urldate = {2020-11-26},
	journal = {arXiv:1903.02891 [cs, stat]},
	author = {Sattler, Felix and Wiedemann, Simon and Müller, Klaus-Robert and Samek, Wojciech},
	month = mar,
	year = {2019},
	note = {arXiv: 1903.02891},
}
\end{document}